\documentclass[conference]{IEEEtran}
\usepackage{times}
\usepackage[utf8]{inputenc}
\usepackage{graphicx}
\usepackage{amsmath}
\usepackage{amssymb}
\usepackage{booktabs}
\usepackage{multirow}
\usepackage{subcaption}

\usepackage{url}
\usepackage{bbm}

\usepackage{fancyhdr}

\usepackage[hidelinks]{hyperref}
\hypersetup{
  colorlinks   = true, 
  urlcolor     = blue, 
  citecolor   = green 
}

\title{ReRankMatch: Semi-Supervised Learning with Semantics-Oriented Similarity Representation}
\author{
\IEEEauthorblockN{Trung Quang Tran}
\IEEEauthorblockA{\textit{School of Computing} \\
\textit{KAIST}\\
Daejeon, South Korea \\
trungtq2019@kaist.ac.kr}
\and
\IEEEauthorblockN{Mingu Kang}
\IEEEauthorblockA{\textit{School of Computing} \\
\textit{KAIST}\\
Daejeon, South Korea \\
mingu.kang@kaist.ac.kr}
\and
\IEEEauthorblockN{Daeyoung Kim}
\IEEEauthorblockA{\textit{School of Computing} \\
\textit{KAIST}\\
Daejeon, South Korea \\
kimd@kaist.ac.kr}
}
\begin{document}

\maketitle

\thispagestyle{fancy}
\pagestyle{fancy}
\fancyhf{}
\lhead{The International Joint Conference on Neural Networks (IJCNN)}
\rhead{July 18-22, 2021}
\cfoot{\thepage}

\begin{abstract}
This paper proposes integrating semantics-oriented similarity representation into RankingMatch, a recently proposed semi-supervised learning method. Our method, dubbed ReRankMatch, aims to deal with the case in which labeled and unlabeled data share non-overlapping categories. ReRankMatch encourages the model to produce the similar image representations for the samples likely belonging to the same class. We evaluate our method on various datasets such as CIFAR-10, CIFAR-100, SVHN, STL-10, and Tiny ImageNet. We obtain promising results (4.21\% error rate on CIFAR-10 with 4000 labels, 22.32\% error rate on CIFAR-100 with 10000 labels, and 2.19\% error rate on SVHN with 1000 labels) when the amount of labeled data is sufficient to learn semantics-oriented similarity representation. The code is made publicly available at \url{https://github.com/tqtrunghnvn/ReRankMatch}.
\end{abstract}

\section{Introduction}

With the burgeoning development of deep neural networks, supervised learning has emerged as a robust approach in a variety of machine learning tasks such as image recognition, voice recognition, or language processing \cite{kolesnikov2019big,xie2020self,clark2017simple}. Supervised learning requires labeled data in training the model. However, obtaining such labeled data comes at a tremendous cost and is time-consuming. In many real-world problems, data labeling demands the knowledge of the experts. For instance, medical data must be analyzed and labeled by doctors.

The big question raised in the community is how to leverage unlabeled data, resulting in unsupervised learning. Unsupervised learning comprises of various research fields such as self-supervised learning \cite{noroozi2016unsupervised,gidaris2018unsupervised}, semi-supervised learning \cite{berthelot2019mixmatch,sohn2020fixmatch}, or metric learning \cite{schroff2015facenet,hermans2017defense}. While self-supervised learning designs pretext tasks to train the model on unlabeled data, semi-supervised learning utilizes both labeled and unlabeled data in a single training process. In another way, metric learning learns the similarity among inputs to cluster the same-class samples together.

Different unsupervised learning approaches could be combined to obtain a more robust method. RankingMatch \cite{anonymous2021rankingmatch}, a current state-of-the-art method, proposed unifying semi-supervised learning and metric learning to build a more robust semi-supervised learning method. The main idea of RankingMatch is leveraging metric learning to encourage the model to produce the similar outputs for not only the different perturbations of the same input but also the same-class inputs. The model output in RankingMatch \cite{anonymous2021rankingmatch} is a vector where each element represents the score of the corresponding class. Intuitively, the dimension of the model output is equal to the number of classes. RankingMatch and many semi-supervised learning methods are designed to deal with unlabeled data sharing overlapping categories with labeled data. For example, if labeled data consists of three classes (\textit{cat}, \textit{dog}, and \textit{bird}), unlabeled data should belong to one of these three classes.

\begin{figure*}[h!]
\centering
\includegraphics[width=7in]{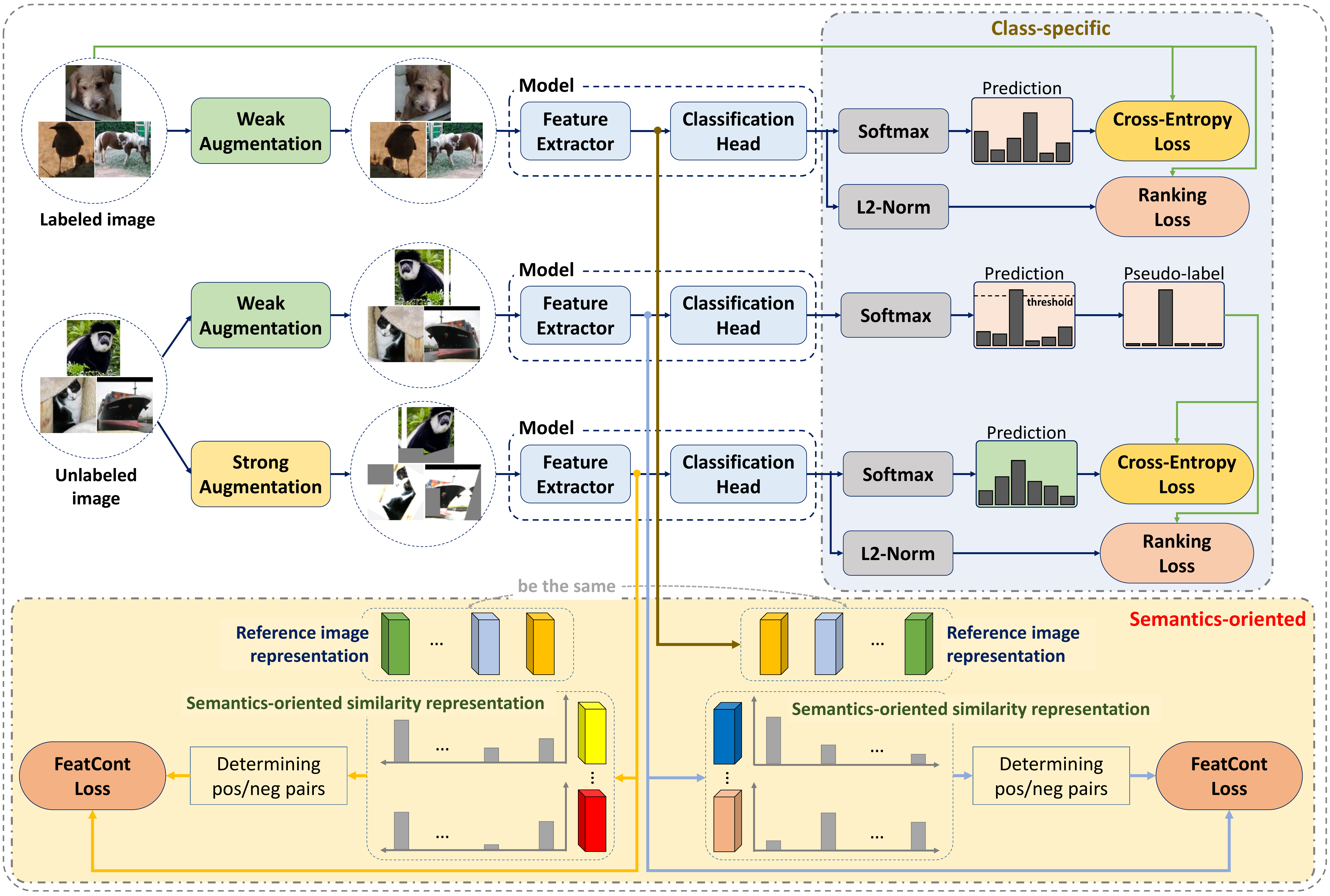}
\caption{Overall framework of ReRankMatch. In addition to class-specific part, we leverage the semantics-oriented similarity representation, expecting that the image representations from the same class could be as similar as possible while the image representations from the different classes should be different.}
\label{fig_framework}
\end{figure*}

However, in reality, unlabeled data commonly do not share overlapping categories with labeled data. Inspired by the success of \cite{chen2020learning} in representation learning in which labeled and unlabeled data share non-overlapping categories, we propose integrating semantics-oriented similarity representation into RankingMatch to address the image classification task. We call our method ReRankMatch, and the overall framework is illustrated in Figure \ref{fig_framework}. The model consists of two main parts: feature extractor and classification head. Given an image, the feature extractor is responsible for understanding the image and outputs the image representation. The image representation is then fed into the classification head to produce the model output called ``logits" score (the scores for the classes).

As shown in Figure \ref{fig_framework}, class-specific part utilizes the ``logits" score to compute Cross-Entropy and Ranking loss, which is similar to RankingMatch \cite{anonymous2021rankingmatch}. Our main work is semantics-oriented part, which leverages the image representation to calculate the feature contrastive loss. Semantics-oriented part aims to encourage the model to leverage unlabeled data without considering category overlapping. Concretely, given a batch of labeled samples, we randomly pick one image representation for each class to become a reference image representation. These reference image representations are then used to compute semantics-oriented similarity representation for unlabeled data. Given an unlabeled image whose the image representation $f_b^U$, the semantics-oriented similarity representation is a vector in which each element is the similarity score between $f_b^U$ and the corresponding reference image representation. Note that the semantics-oriented similarity representation could be computed even when labeled and unlabeled data share non-overlapping categories. From the computed semantics-oriented similarity representation, we can assign pseudo positive/negative labels for unlabeled data and then compute the feature contrastive loss, which will be presented in Section \ref{sec_method}. We summarize our key contributions as follows:
\begin{itemize}
    \item We propose ReRankMatch, a semi-supervised learning (SSL) method leveraging the semantics-oriented similarity representation to deal with the case where labeled and unlabeled data share non-overlapping categories.
    \item We conduct experiments on various standard SSL benchmarks such as CIFAR-10, CIFAR-100, SVHN, and STL-10. We also evaluate our method on a larger dataset, Tiny ImageNet.
    \item Our extensive experimental results show that semantics-oriented similarity representation becomes meaningful and helpful if the amount of labeled data is sufficient to learn semantics-oriented similarity representation.
\end{itemize}

\section{Related Work}

This section reviews two important contents, which are directly relevant to our work.

\subsection{RankingMatch}

RankingMatch \cite{anonymous2021rankingmatch} is a recently proposed semi-supervised learning (SSL) method, achieving state-of-the-art results across many standard SSL benchmarks. If the semantics-oriented part in Figure \ref{fig_framework} is removed, the remaining ones constitute the RankingMatch's diagram. RankingMatch is an integration of consistency regularization SSL approach and metric learning:
\begin{itemize}
    \item \textbf{Consistency regularization} is one of the most successful SSL approaches. The given input is perturbed in different ways. Consistency regularization encourages the model to produce unchanged with different perturbations of the same input.
    \item \textbf{Metric learning} is a research field that does not directly assign semantic labels for given inputs but aims at measuring the similarity among the inputs. That is, metric learning tries to make the model outputs of the same-class samples as similar as possible.
\end{itemize}

RankingMatch argued that consistency regularization is not enough, and the model should also tend to produce the similar outputs for the samples from the same class. By leveraging both consistency regularization and metric learning, RankingMatch encourages the model to produce the similar outputs for not only the different perturbations of the same sample but also the same-class samples. Concretely, in addition to Cross-Entropy loss for image classification, RankingMatch utilized Ranking loss of metric learning, as illustrated in the class-specific part in Figure \ref{fig_framework}. Two types of Ranking loss used in RankingMatch are Tripet and Contrastive loss. RankingMatch also introduced a new version of Ranking loss, which is BatchMean Triplet loss. We use BatchMean Triplet and Contrastive loss in our ReRankMatch.

\subsection{Semantics-Oriented Similarity Representation}

Semantics-oriented similarity representation was successfully exploited by \cite{chen2020learning} for person re-identification and image retrieval. \cite{chen2020learning} proposed a novel meta-learning scheme, consisting of two learning phases, to utilize both labeled and unlabeled data. While the first phase uses only labeled data, the second phase utilizes both labeled and unlabeled data.

In the first phase, the labeled set is split into a meta-training set $M_T$ and a meta-validation set $M_V$, where $M_T$ and $M_V$ share non-overlapping labels. Let $C_{M_T}$ denote the number of classes in $M_T$. \cite{chen2020learning} samples $C_{M_T}$ reference images from $M_T$ in which each reference image is for each class. The model then extracts the feature $\hat{f}^{M_T}$ for each reference image $\hat{x}^{M_T}$. With each feature $f^{M_V}$ extracted from $M_V$, \cite{chen2020learning} computes the class-wise similarity scores between $f^{M_V}$ and all reference features $\hat{f}^{M_T}$, and these scores constitute a semantics-oriented similarity representation. The semantics-oriented similarity representations are then used to compute the similarity contrastive loss.

In the second phase, the reference images are chosen from the entire labeled set. For each unlabeled sample, \cite{chen2020learning} computes the semantics-oriented similarity representations as similar to the first phase. However, these semantics-oriented similarity representations are used to assign pseudo positive/negative labels for unlabeled data, allowing computing the feature contrastive loss as done on labeled data. Since $M_T$ and $M_V$ in the first phase share non-overlapping labels, the second phase could work even when labeled and unlabeled data share non-overlapping categories.

\section{ReRankMatch}
\label{sec_method}

ReRankMatch consists of two main parts: class-specific and semantics-oriented. This section starts to describe the overall framework of ReRankMatch. Then, the class-specific and semantics-oriented part will be presented in detail.

\subsection{Overall Framework}

The overall framework of ReRankMatch is shown in Figure \ref{fig_framework}. While labeled data only uses weak augmentation, unlabeled data uses both weak and strong augmentation. Standard padding-and-cropping and horizontal flipping strategies are used for weak augmentation, whereas strong augmentation additionally utilizes more advanced transformations such as RandAugment \cite{cubuk2020randaugment} and Cutout \cite{devries2017improved}. As shown in Figure \ref{fig_framework}, the model consists of two main parts: feature extractor and classification head. The output of the feature extractor is called image representation, and the output of the classification head is ``logits" score.

For weakly-augmented labeled data, the image representations are used to sample reference image representations, and the ``logits" scores are for computing Cross-Entropy and Ranking loss. For unlabeled data, the ``logits" scores corresponding to the weakly-augmented versions serve to compute pseudo-labels. These pseudo-labels then become the target labels in computing Cross-Entropy and Ranking loss when the strongly-augmented unlabeled data is fed into the model. Finally, the image representations corresponding to weakly-augmented and strongly-augmented unlabeled data are used along with the sampled reference image representations to compute the feature contrastive loss in the semantics-oriented part.

Let $\mathcal{X}=\{(f_b^\mathcal{X},p_b^\mathcal{X},l_b):b\in(1,...,B)\}$ be a batch of $B$ weakly-augmented labeled samples, where $f_b^\mathcal{X}$ and $p_b^\mathcal{X}$ are the image representation and ``logits" score respectively, and $l_b$ is the corresponding one-hot label. Let $\mathcal{U}=\{f_b^{\mathcal{U}_w},p_b^{\mathcal{U}_w},f_b^{\mathcal{U}_s},p_b^{\mathcal{U}_s}:b\in(1,...,\mu B)\}$ define a batch of $\mu B$ unlabeled samples with a coefficient $\mu$ determining the relative size of $\mathcal{X}$ and $\mathcal{U}$. $f_b^{\mathcal{U}_w}$ and $p_b^{\mathcal{U}_w}$ are the image representation and ``logits" score corresponding to weakly-augmented unlabeled data, respectively; $f_b^{\mathcal{U}_s}$ and $p_b^{\mathcal{U}_s}$ are the image representation and ``logits" score corresponding to strongly-augmented unlabeled data, respectively. Finally, let $\mathrm{H}(v,q)$ be Cross-Entropy loss over the predicted class distribution $q$ and the target label $v$.

The objective function of ReRankMatch comprises of three types of loss, formulated as follows:
\begin{equation}
    \mathcal{L}^{ReRankMatch}=\mathcal{L}^{CE} + \lambda_r \mathcal{L}^{Rank} + \lambda_s \mathcal{L}^{FeatCont}
    \label{equation_totalloss}
\end{equation}
where $\mathcal{L}^{CE}$, $\mathcal{L}^{Rank}$, and $\mathcal{L}^{FeatCont}$ are Cross-Entropy, Ranking, and feature contrastive loss, respectively; $\lambda_r$ and $\lambda_s$ are loss weights. $\mathcal{L}^{CE}$ and $\mathcal{L}^{Rank}$ will be presented in Section \ref{sec_class_specific}, and $\mathcal{L}^{FeatCont}$ will be elaborated in Section \ref{sec_semantics_oriented}.

\subsection{Class-Specific}
\label{sec_class_specific}

\subsubsection{Cross-Entropy Loss}

For labeled data, the standard Cross-Entropy loss could be simply computed by:
\begin{equation}
    \mathcal{L}_\mathcal{X}^{CE}=\frac{1}{B}\sum\limits_{b=1}^B \mathrm{H}(l_b,\mathrm{Softmax}(p_b^\mathcal{X}))
\end{equation}

\noindent
For unlabeled data, let $\tilde{q}_b$ denote the result when applying a softmax function to $p_b^{\mathcal{U}_w}$: $\tilde{q}_b=\mathrm{Softmax}(p_b^{\mathcal{U}_w})$. Let $\hat{q}_b$ be the pseudo-label: $\hat{q}_b=\mathrm{argmax}(\tilde{q}_b)$, which corresponds to the class having the highest probability. Notably, $\mathrm{argmax}$ is assumed to produce the valid one-hot pseudo-label for simplicity. Cross-Entropy loss when feeding the strongly-augmented unlabeled data into the model is computed as follows:
\begin{equation}
    \mathcal{L}_\mathcal{U}^{CE}=\frac{1}{\mu B}\sum\limits_{b=1}^{\mu B}\mathbbm{1}(\mathrm{max}(\tilde{q}_b)\geq \tau)\;\mathrm{H}(\hat{q}_b,\mathrm{Softmax}(p_b^{\mathcal{U}_s}))
    \label{equation_luce}
\end{equation}
where $\tau$ is a confidence threshold to ignore low-confidence predictions. Finally, Cross-Entropy loss for both labeled and unlabeled data is computed as:
\begin{equation}
    \mathcal{L}^{CE} = \mathcal{L}_\mathcal{X}^{CE} + \lambda_u \mathcal{L}_\mathcal{U}^{CE}
\end{equation}
where $\lambda_u$ is the loss weight.

\subsubsection{Ranking Loss}

As shown in Figure \ref{fig_framework}, an $L_2$-normalization is applied to the ``logits" scores before computing Ranking loss. Therefore, let $\mathcal{C}^\mathcal{X}$ define a batch of $L_2$-normalized ``logits" scores corresponding to weakly-augmented labeled data and $y_i$ be the ground-truth label of the $L_2$-normalized ``logits" score $i$; let $\mathcal{C}^{\mathcal{U}}$ define a batch of $L_2$-normalized ``logits" scores corresponding to strongly-augmented unlabeled data and $q_i$ be the pseudo-label of the $L_2$-normalized ``logits" score $i$. Finally, let $a$, $p$, and $n$ denote the anchor, positive, and negative sample, respectively. $a$ and $p$ represent the $L_2$-normalized ``logits" scores having the same label, while $a$ and $n$ are for the $L_2$-normalized ``logits" scores having the different labels. There are two types of Ranking loss used in ReRankMatch: BatchMean Triplet loss ($\mathcal{L}_{BM}^{Rank}$) and Contrastive loss ($\mathcal{L}_{CT}^{Rank}$). Note that $\mathcal{L}^{Rank}$ in Equation \ref{equation_totalloss} could be either $\mathcal{L}_{BM}^{Rank}$ or $\mathcal{L}_{CT}^{Rank}$. Referring to RankingMatch \cite{anonymous2021rankingmatch}, the loss functions are defined as follows.

\noindent
\textbf{\textit{BatchMean Triplet loss}}:
\begin{equation}
\small
\begin{gathered}
    \mathcal{L}_\mathcal{X}^{BM} = \frac{1}{|\mathcal{C}^\mathcal{X}|}\sum\limits_{a\in \mathcal{C}^\mathcal{X}} f(m + \frac{1}{|\mathcal{C}^\mathcal{X}|}\sum\limits_{\substack{p\in \mathcal{C}^\mathcal{X} \\ y_p=y_a}}d_{a,p} - \frac{1}{|\mathcal{C}^\mathcal{X}|}\sum\limits_{\substack{n\in \mathcal{C}^\mathcal{X} \\ y_n\neq y_a}}d_{a,n}) \\
    \mathcal{L}_\mathcal{U}^{BM} = \frac{1}{|\mathcal{C}^\mathcal{U}|}\sum\limits_{a\in \mathcal{C}^\mathcal{U}} f(m + \frac{1}{|\mathcal{C}^\mathcal{U}|}\sum\limits_{\substack{p\in \mathcal{C}^\mathcal{U} \\ q_p=q_a}}d_{a,p} - \frac{1}{|\mathcal{C}^\mathcal{U}|}\sum\limits_{\substack{n\in \mathcal{C}^\mathcal{U} \\ q_n\neq q_a}}d_{a,n})\\
    \mathcal{L}_{BM}^{Rank} = \mathcal{L}_\mathcal{X}^{BM} + \mathcal{L}_\mathcal{U}^{BM}
    \label{equation_BM}
\end{gathered}
\end{equation}
where $d_{i,j}$ is the Euclidean distance between two $L_2$-normalized ``logits" scores $i$ and $j$, and $m$ is the margin. $f(\bullet)$ is the function to avoid revising ``already correct" triplets. We use a softplus function ($f(\bullet)=\ln{(1+\exp{(\bullet)})}$) as in RankingMatch.

\noindent
\textbf{\textit{Contrastive loss}}:
\begin{equation}
\small
\begin{gathered}
    \mathcal{L}_\mathcal{X}^{CT}=\frac{1}{N}\sum\limits_{\substack{a,p\in \mathcal{C}^\mathcal{X} \\ a\neq p \\ y_a=y_p}}-\ln{\frac{\exp{(sim_{a,p}/T)}}{\exp{(sim_{a,p}/T)} + \sum\limits_{\substack{n\in \mathcal{C}^\mathcal{X} \\ y_n\neq y_a}}\exp{(sim_{a,n}/T)}}} \\
    \mathcal{L}_\mathcal{U}^{CT}=\frac{1}{N}\sum\limits_{\substack{a,p\in \mathcal{C}^\mathcal{U} \\ a\neq p \\ q_a=q_p}}-\ln{\frac{\exp{(sim_{a,p}/T)}}{\exp{(sim_{a,p}/T)} + \sum\limits_{\substack{n\in \mathcal{C}^\mathcal{U} \\ q_n\neq q_a}}\exp{(sim_{a,n}/T)}}} \\
    \mathcal{L}_{CT}^{Rank} = \mathcal{L}_\mathcal{X}^{CT} + \mathcal{L}_\mathcal{U}^{CT}
    \label{equation_Con}
\end{gathered}
\end{equation}
where $sim_{i,j}$ is the cosine similarity between two $L_2$-normalized ``logits" scores $i$ and $j$, and $T$ is the temperature parameter.

\subsection{Semantics-Oriented}
\label{sec_semantics_oriented}

The semantics-oriented part uses the image representations to compute the feature contrastive loss $\mathcal{L}^{FeatCont}$, consisting of four steps as follows.

\noindent
\textbf{\textit{Step 1: Choosing reference image representations from labeled set}}

For each category appearing in $\mathcal{X}$, we randomly sample a corresponding image representation to become a reference image representation $\hat{f}^\mathcal{X}$. The number of reference image representations is equal to the number of unique labels in $\mathcal{X}$. For instance, we have seven image representations with their labels: $\{(f_1^\mathcal{X},l_1=1),(f_2^\mathcal{X},l_2=3),(f_3^\mathcal{X},l_3=1),(f_4^\mathcal{X},l_4=1),(f_5^\mathcal{X},l_5=4),(f_6^\mathcal{X},l_6=3),(f_7^\mathcal{X},l_7=3)\}$. As a result, we have three unique labels which are $1$, $3$, and $4$. By picking the image representation for each unique label randomly, we obtain three reference image representations $\{f_3^\mathcal{X},f_6^\mathcal{X},f_5^\mathcal{X}\}$.

\noindent
\textbf{\textit{Step 2: Computing semantics-oriented similarity representation for unlabeled data}}

This step is the same for weakly-augmented and strongly-augmented unlabeled data, so we do not explicitly mention weak or strong augmentation for a convenient explanation. To obtain the semantics-oriented similarity representation $s_b^\mathcal{U}$, we compute the class-wise similarity scores between $f_b^\mathcal{U}$ and all reference image representations sampled in the previous step. $s_b^\mathcal{U}$ has the dimension being equal to the number of reference image representations. The $k^{th}$ element of $s_b^\mathcal{U}$ is defined as:
\begin{equation}
    s_b^\mathcal{U}(k) = \mathrm{sim}(f_b^\mathcal{U},\hat{f}_k^\mathcal{X})
\end{equation}
where $\hat{f}_k^\mathcal{X}$ is the $k^{th}$ reference image representation, and we use cosine similarity for the ``sim" function.

\noindent
\textbf{\textit{Step 3: Assigning pseudo positive/negative labels for unlabeled data}}

Similar to Step 2, we do not explicitly mention weak and strong augmentation because of the same procedure for weakly-augmented and strongly-augmented unlabeled data. Given two semantics-oriented similarity representations $s_i^\mathcal{U}$ and $s_j^\mathcal{U}$, we can determine whether the two corresponding images come from the same class ($t=1$) or not ($t=0$). Concretely,
\begin{equation}
    t = 
    \left\{
        \begin{array}{ll}
            1, & \mbox{if } ||s_i^\mathcal{U} - s_j^\mathcal{U}|| < \psi\\
            0, & \mbox{otherwise}
        \end{array}
    \right.
\end{equation}
with $\psi > 0$ is a threshold.

\noindent
\textbf{\textit{Step 4: Computing the feature contrastive loss}}

Given a pair of two image representations corresponding to weakly-augmented unlabeled data ($f_i^{\mathcal{U}_w}$ and $f_j^{\mathcal{U}_w}$) and a pair of two image representations corresponding to strongly-augmented unlabeled data ($f_p^{\mathcal{U}_s}$ and $f_q^{\mathcal{U}_s}$), the feature contrastive loss could be defined as follows:
\begin{equation}
\begin{gathered}
    \mathcal{L}_{\mathcal{U}_w}^{FeatCont}(i,j) = \;\;\;\;\;\;\;\;\;\;\;\;\;\;\;\;\;\;\;\;\;\;\;\;\;\;\;\;\;\;\;\;\;\;\;\;\;\;\;\;\;\;\;\;\;\;\;\;\;\;\;\;\;\;\;\;\;\;\;\;\;\; \\
    \;\;\;\;\;\;\;\; t\cdot ||f_i^{\mathcal{U}_w} - f_j^{\mathcal{U}_w}|| + (1-t)\cdot \mathrm{max}(0,\phi - ||f_i^{\mathcal{U}_w} - f_j^{\mathcal{U}_w}||) \\
    \mathcal{L}_{\mathcal{U}_s}^{FeatCont}(p,q) = \;\;\;\;\;\;\;\;\;\;\;\;\;\;\;\;\;\;\;\;\;\;\;\;\;\;\;\;\;\;\;\;\;\;\;\;\;\;\;\;\;\;\;\;\;\;\;\;\;\;\;\;\;\;\;\;\;\;\;\;\;\; \\
    \;\;\;\;\;\;\;\;\;\;\; t\cdot ||f_p^{\mathcal{U}_s} - f_q^{\mathcal{U}_s}|| + (1-t)\cdot \mathrm{max}(0,\phi - ||f_p^{\mathcal{U}_s} - f_q^{\mathcal{U}_s}||)
\end{gathered}
\end{equation}
where $\phi > 0$ denotes the margin.

$\mathcal{L}_{\mathcal{U}_w}^{FeatCont}$ and $\mathcal{L}_{\mathcal{U}_s}^{FeatCont}$ will be computed over all pairs of image representations on weakly-augmented and strongly-augmented unlabeled data, respectively. Finally, $\mathcal{L}^{FeatCont}$ is the summation of $\mathcal{L}_{\mathcal{U}_w}^{FeatCont}$ and $\mathcal{L}_{\mathcal{U}_s}^{FeatCont}$:
\begin{equation}
    \mathcal{L}^{FeatCont} = \mathcal{L}_{\mathcal{U}_w}^{FeatCont} + \mathcal{L}_{\mathcal{U}_s}^{FeatCont}
\end{equation}
By minimizing $\mathcal{L}^{FeatCont}$, we expect that the image representations from the same class could be as similar as possible while the image representations from the different classes should be different by at least a margin $\phi$.

\section{Experiments}

We compare our method against various semi-supervised learning (SSL) approaches such as MixMatch \cite{berthelot2019mixmatch}, RealMix \cite{nair2019realmix}, ReMixMatch \cite{berthelot2019remixmatch}, FixMatch \cite{sohn2020fixmatch}, and RankingMatch \cite{anonymous2021rankingmatch}. We conduct experiments on standard SSL benchmarks such as CIFAR-10 \cite{krizhevsky2009learning}, CIFAR-100 \cite{krizhevsky2009learning}, SVHN \cite{netzer2011reading}, and STL-10 \cite{coates2011analysis}. We also verify the performance of our method on a larger dataset, which is Tiny ImageNet\footnote{Stanford University. \url{http://cs231n.stanford.edu/}}.

\subsection{Implementation Details}

We utilize Wide ResNet architecture \cite{zagoruyko2016wide} in our experiments. An SGD optimizer with momentum is adopted in training the models. We apply a cosine learning rate decay \cite{loshchilov2016sgdr}, which effectively decays the learning rate using a cosine curve. We use a same set of hyperparameters ($B = 64$, $\mu = 7$, $\tau = 0.95$, $m = 0.5$, $T = 0.2$, $\psi = 0.5$, $\phi = 0.3$, $\lambda_u = 1$, $\lambda_r = 1$, and $\lambda_s = 1$) across all datasets and all amounts of labeled samples except that a batch size of 32 ($B = 32$) is used for the STL-10 dataset. In all our experiments, $^\text{BM}$ and $^\text{CT}$ denote the method using BatchMean Triplet loss and Contrastive loss respectively, and FixMatch$^{\text{RA}}$ means FixMatch with using RandAugment.

\subsection{CIFAR-10}
\label{sec_cifar10}

CIFAR-10 \cite{krizhevsky2009learning} is a widely used dataset that consists of $32\times 32$ color images with 10 classes. There are 50000 training images and 10000 test images. Following standard practice, as mentioned by \cite{oliver2018realistic}, we divide training images into train and validation split, with 45000 images for training and 5000 images for validation. In the train split, we discard all except a number of labels (40, 250, and 4000 labels) to vary the labeled data set size. We use Wide ResNet-28-2 architecture with 1.5 million parameters, and our models are trained for 256 epochs. The results are shown in Table \ref{table_cifar10}.

\begin{table}[h!]
\centering
\scalebox{1.1}{
\begin{tabular}{lrrr}
\toprule
Method            & 40 labels & 250 labels & 4000 labels \\
\midrule
MixMatch          & -         & 11.08$\pm$0.87  & 6.24$\pm$0.06   \\
RealMix           & -         & 9.79$\pm$0.75  & 6.39$\pm$0.27   \\
ReMixMatch        & -         & 6.27$\pm$0.34  & 5.14$\pm$0.04   \\
FixMatch$^{\text{RA}}$     & 13.81$\pm$3.37 & 5.07$\pm$0.65  & 4.26$\pm$0.05   \\
RankingMatch$^{\text{BM}}$ & \textbf{13.43}$\pm$2.33       & \textbf{4.87}$\pm$0.08        & 4.29$\pm$0.03         \\
RankingMatch$^{\text{CT}}$ & 14.98$\pm$3.06       & 5.13$\pm$0.02        & 4.32$\pm$0.12         \\
\midrule
ReRankMatch$^{\text{BM}}$ & 66.77$\pm$3.91       & 20.07$\pm$1.16        & \textbf{4.21}$\pm$0.07         \\
ReRankMatch$^{\text{CT}}$ & 18.25$\pm$9.44       & 6.02$\pm$1.31        & 4.40$\pm$0.06         \\
\bottomrule
\end{tabular}
}
\caption{Error rate (\%) on CIFAR-10.}
\label{table_cifar10}
\end{table}

As shown in Table \ref{table_cifar10}, semantics-oriented similarity representation has strong effect on RankingMatch's performance, especially with a small number of labels. With 250 and 40 labels, ReRankMatch$^{\text{BM}}$ significantly decreases performance compared to RankingMatch$^{\text{BM}}$. Especially, the results also show that ReRankMatch$^{\text{CT}}$ outperforms ReRankMatch$^{\text{BM}}$ when the number of labeled data is small. For instance, ReRankMatch$^{\text{CT}}$ reduces the error rate by 14.05\% and 48.52\% compared to ReRankMatch$^{\text{BM}}$ with 250 and 40 labels, respectively. This could be explained by the fact that ReRankMatch$^{\text{CT}}$ uses cosine similarity for both class-specific and semantics-oriented part, so ReRankMatch$^{\text{CT}}$ might be more predominant than ReRankMatch$^{\text{BM}}$.

However, ReRankMatch can obtain a slightly better performance than RankingMatch with more labels. \textbf{Specifically}, ReRankMatch$^{\text{BM}}$ sets a new state-of-the-art result, 4.21\% error rate, with 4000 labels.

\subsection{CIFAR-100}
\label{sec_cifar100}

CIFAR-100 \cite{krizhevsky2009learning} contains 100 classes with 50000 training images and 10000 test images. Similar to CIFAR-10, we also divide training images into train and validation split, with 45000 images for training and 5000 images for validation. In the train split, we discard all except a number of labels (400, 2500, and 10000 labels) to vary the labeled data set size. We use Wide ResNet-28-2-Large architecture with 26 million parameters and train the models for 128 epochs. The results are shown in Table \ref{table_cifar100}.

\begin{table}[h!]
\centering
\scalebox{1.1}{
\begin{tabular}{lrrr}
\toprule
Method            & 400 labels & 2500 labels & 10000 labels \\
\midrule
MixMatch          & - & - & 25.88$\pm$0.30   \\
FixMatch$^{\text{RA}}$     & \textbf{48.85}$\pm$1.75 & \textbf{28.29}$\pm$0.11  & 22.60$\pm$0.12   \\
RankingMatch$^{\text{BM}}$ & 49.57$\pm$0.67       & 29.68$\pm$0.60        & 23.18$\pm$0.03         \\
RankingMatch$^{\text{CT}}$ & 56.90$\pm$1.47       & 28.39$\pm$0.67        & 22.35$\pm$0.10         \\
\midrule
ReRankMatch$^{\text{BM}}$ & 87.46$\pm$3.65       & 49.71$\pm$0.46        & 25.38$\pm$0.93         \\
ReRankMatch$^{\text{CT}}$ & 69.62$\pm$1.33       & 31.75$\pm$0.33        & \textbf{22.32}$\pm$0.65         \\
\bottomrule
\end{tabular}
}
\caption{Error rate (\%) on CIFAR-100.}
\label{table_cifar100}
\end{table}

As shown in Table \ref{table_cifar100}, the tendency of ReRankMatch's models is relatively similar to the case of CIFAR-10. ReRankMatch significantly decreases performance compared to RankingMatch, with a small number of labeled data (2500 and 400 labels). ReRankMatch$^{\text{CT}}$ is superior to ReRankMatch$^{\text{BM}}$ in all cases. Concretely, ReRankMatch$^{\text{CT}}$ reduces the error rate by 3.06\%, 17.96\%, and 17.84\% compared to ReRankMatch$^{\text{BM}}$ in the case of 10000, 2500, and 400 labels respectively. \textbf{Importantly}, ReRankMatch$^{\text{CT}}$ achieves a new state-of-the-art result, 22.32\% error rate, with 10000 labels.

\subsection{SVHN}
\label{sec_svhn}

SVHN \cite{netzer2011reading}, Street View House Numbers, is a real-world dataset that consists of 73257 training images and 26032 test images. The resolution of images in the SVHN dataset is $32\times 32$. We divide training images into train split with 65937 images and validation split with 7320 images. In the train split, we discard all except a number of labels (40, 250, and 1000 labels) to vary the labeled data set size. We use Wide ResNet-28-2 architecture with 1.5 million parameters, and our models are trained for 128 epochs. The results are shown in Table \ref{table_svhn}.

\begin{table}[h!]
\centering
\scalebox{1.1}{
\begin{tabular}{lrrr}
\toprule
Method            & 40 labels & 250 labels & 1000 labels \\
\midrule
MixMatch          & -         & 3.78$\pm$0.26  & 3.27$\pm$0.31   \\
RealMix           & -         & 3.53$\pm$0.38  & -   \\
ReMixMatch        & -         & 3.10$\pm$0.50  & 2.83$\pm$0.30   \\
FixMatch$^{\text{RA}}$     & \textbf{3.96}$\pm$2.17 & 2.48$\pm$0.38  & 2.28$\pm$0.11   \\
RankingMatch$^{\text{BM}}$ & \underline{21.02}$\pm$8.06       & \textbf{2.24}$\pm$0.07        & 2.32$\pm$0.07         \\
RankingMatch$^{\text{CT}}$ & 27.20$\pm$2.90       & 2.33$\pm$0.06        & 2.23$\pm$0.11         \\
\midrule
ReRankMatch$^{\text{BM}}$ & 74.57$\pm$3.95       & 7.76$\pm$1.19        & 2.32$\pm$0.08         \\
ReRankMatch$^{\text{CT}}$ & \underline{20.25}$\pm$4.43       & 2.44$\pm$0.07        & \textbf{2.19}$\pm$0.09         \\
\bottomrule
\end{tabular}
}
\caption{Error rate (\%) on SVHN.}
\label{table_svhn}
\end{table}

As shown in Table \ref{table_svhn}, overall, the arguments are similar to those in the case of CIFAR-10 and CIFAR-100. ReRankMatch$^{\text{CT}}$ is still better than ReRankMatch$^{\text{BM}}$ in all cases, which are 0.13\%, 5.32\%, and 54.32\% error rate improvement with 1000, 250, and 40 labels respectively.

The results also show that ReRankMatch decreases performance compared to RankingMatch when the number of labels is small. However, there is a case where ReRankMatch is slightly better than RankingMatch. That is, ReRankMatch$^{\text{CT}}$ reduces the error rate by 0.77\% compared to RankingMatch$^{\text{BM}}$ with 40 labels. \textbf{Especially}, ReRankMatch$^{\text{CT}}$ sets a new state-of-the-art result, 2.19\% error rate, with 1000 labels.

\subsection{STL-10}
\label{sec_stl10}

STL-10 \cite{coates2011analysis} is a dataset designed for unsupervised learning, containing 5000 labeled images and 100000 unlabeled images. We use Wide ResNet-37-2 architecture with 5.9 million parameters, and our models are trained for 128 epochs. STL-10 consists of ten pre-defined folds with 1000 labeled images each. Given a fold, we use 4000 other labeled images out of 5000 labeled training images as the validation split. STL-10 has a test set with 8000 labeled images. Table \ref{table_stl10} shows the results on 1000-label splits. The results of SWWAE and CC-GAN are cited from \cite{zhao2015stacked} and \cite{denton2016semi}, respectively. 

\begin{table}[h!]
\centering
\scalebox{1.1}{
\begin{tabular}{lrrr}
\toprule
Method            & Error Rate \\
\midrule
SWWAE     & 25.67 \\
CC-GAN    & 22.21 \\
MixMatch          & 10.18$\pm$1.46 \\
ReMixMatch        & 6.18$\pm$1.24 \\
FixMatch$^{\text{RA}}$     & 7.98$\pm$1.50 \\
RankingMatch$^{\text{BM}}$ & \textbf{5.96}$\pm$0.07 \\
RankingMatch$^{\text{CT}}$ & 7.55$\pm$0.37 \\
\midrule
ReRankMatch$^{\text{BM}}$ & 7.11$\pm$0.13 \\
ReRankMatch$^{\text{CT}}$ & 6.96$\pm$0.29 \\
\bottomrule
\end{tabular}
}
\caption{Error rate (\%) on STL-10.}
\label{table_stl10}
\end{table}

As shown in Table \ref{table_stl10}, ReRankMatch outperforms SWWAE, CC-GAN, and MixMatch. We achieve competitive results compared to FixMatch and RankingMatch with the same data augmentation (RandAugment). ReRankMatch$^{\text{BM}}$ and ReRankMatch$^{\text{CT}}$ reduce the error rate by 0.87\% and 1.02\% compared to FixMatch$^{\text{RA}}$, respectively. In comparison with RankingMatch, ReRankMatch$^{\text{BM}}$ is worse than RankingMatch$^{\text{BM}}$, but ReRankMatch$^{\text{CT}}$ reduces the error rate by 0.59\% compared to RankingMatch$^{\text{CT}}$.

\subsection{Tiny ImageNet}
\label{sec_tinyimagenet}

Tiny ImageNet is a compact version of ImageNet, including 100000 training images, 10000 validation images, and 10000 test images. Tiny ImageNet consists of 200 classes and $64\times 64$ images. The ground-truth labels of test images are not available, so we use 10000 validation images as the test set to evaluate our method. We divide 100000 training images into 90000 and 10000 images for train and validation split, respectively. For the semi-supervised learning setting, we use 10\% of train split as labeled data and treat the rest as unlabeled data. As a result, there are 9000 labeled images and 81000 unlabeled images. Note that this setting is the same with other methods, so our results are comparable. We use Wide ResNet-37-2 architecture with 5.9 million parameters, and our models are trained for 128 epochs. The results are shown in Table \ref{table_tinyimagenet}.

\begin{table}[h!]
\centering
\scalebox{1.1}{
\begin{tabular}{lrrr}
\toprule
Method            & Error Rate \\
\midrule
MixMatch          & 59.40$\pm$0.47 \\
FixMatch$^{\text{RA}}$     & 52.09$\pm$0.14 \\
RankingMatch$^{\text{BM}}$ & 51.47$\pm$0.25 \\
RankingMatch$^{\text{CT}}$ & \textbf{49.10}$\pm$0.41 \\
\midrule
ReRankMatch$^{\text{BM}}$ & 58.62$\pm$0.57 \\
ReRankMatch$^{\text{CT}}$ & 55.13$\pm$0.52 \\
\bottomrule
\end{tabular}
}
\caption{Error rate (\%) on Tiny ImageNet. The result of MixMatch is produced by us. The result of FixMatch$^{\text{RA}}$ is cited from RankingMatch.}
\label{table_tinyimagenet}
\end{table}

As shown in Table \ref{table_tinyimagenet}, ReRankMatch is superior to MixMatch. ReRankMatch$^{\text{BM}}$ and ReRankMatch$^{\text{CT}}$ reduce the error rate by 0.78\% and 4.27\% compared to MixMatch, respectively. Compared to other methods, ReRankMatch has worse results. The result also shows that ReRankMatch$^{\text{CT}}$ is better than ReRankMatch$^{\text{BM}}$, where ReRankMatch$^{\text{CT}}$ improves the error rate by 3.49\% compared to ReRankMatch$^{\text{BM}}$.

\subsection{Ablation Study}

\subsubsection{Loss Weight for Feature Contrastive Loss}

We conduct an ablation study with different values of the loss weight $\lambda_s$, as shown in Table \ref{table_loss_weight}. The loss weight $\lambda_s$ does not affect the performance of ReRankMatch, and the results are similar among different values of $\lambda_s$.

\begin{table}[h!]
\centering
\scalebox{1.1}{
\begin{tabular}{lrr}
\toprule
Ablation            & \shortstack[r]{CIFAR-10\\4000 labels} & \shortstack[r]{SVHN\\1000 labels} \\
\midrule
ReRankMatch$^{\text{CT}}$ ($\lambda_s = 0.5$) & 95.32 & 97.77 \\
ReRankMatch$^{\text{CT}}$ ($\lambda_s = 1.0$) & 95.53 & 97.69 \\
ReRankMatch$^{\text{CT}}$ ($\lambda_s = 2.0$) & 95.60 & 97.78 \\
\bottomrule
\end{tabular}
}
\caption{Ablation study with loss weight $\lambda_s$. All results are top-1 accuracy (\%).}
\label{table_loss_weight}
\end{table}

\begin{figure*}[h!]
\centering
\subfloat[ReRankMatch$^{\text{BM}}$ on CIFAR-10]{\includegraphics[width=2.95in]{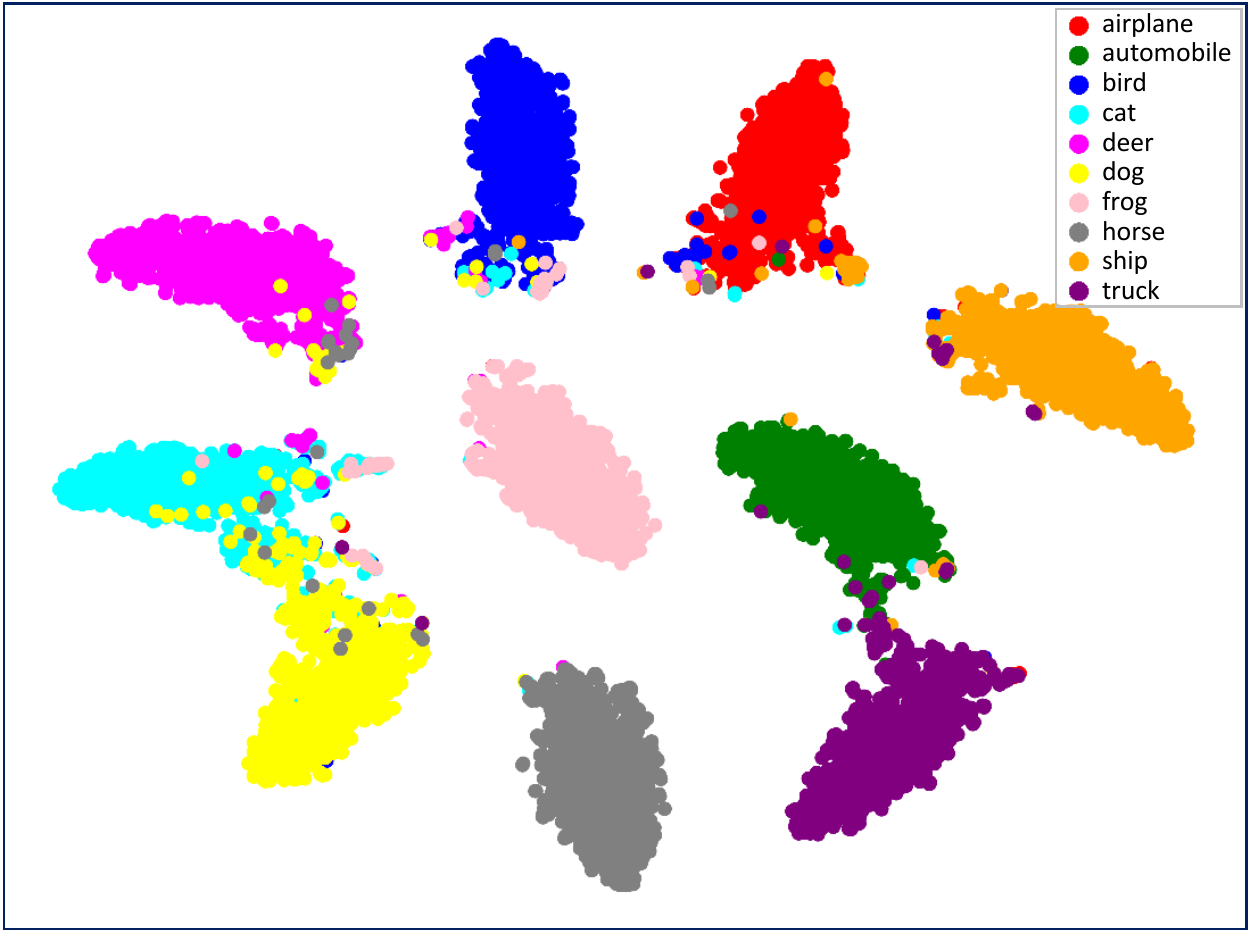}
\label{fig_main_methods_mixmatch}}
\hfil
\subfloat[ReRankMatch$^{\text{CT}}$ on CIFAR-10]{\includegraphics[width=2.95in]{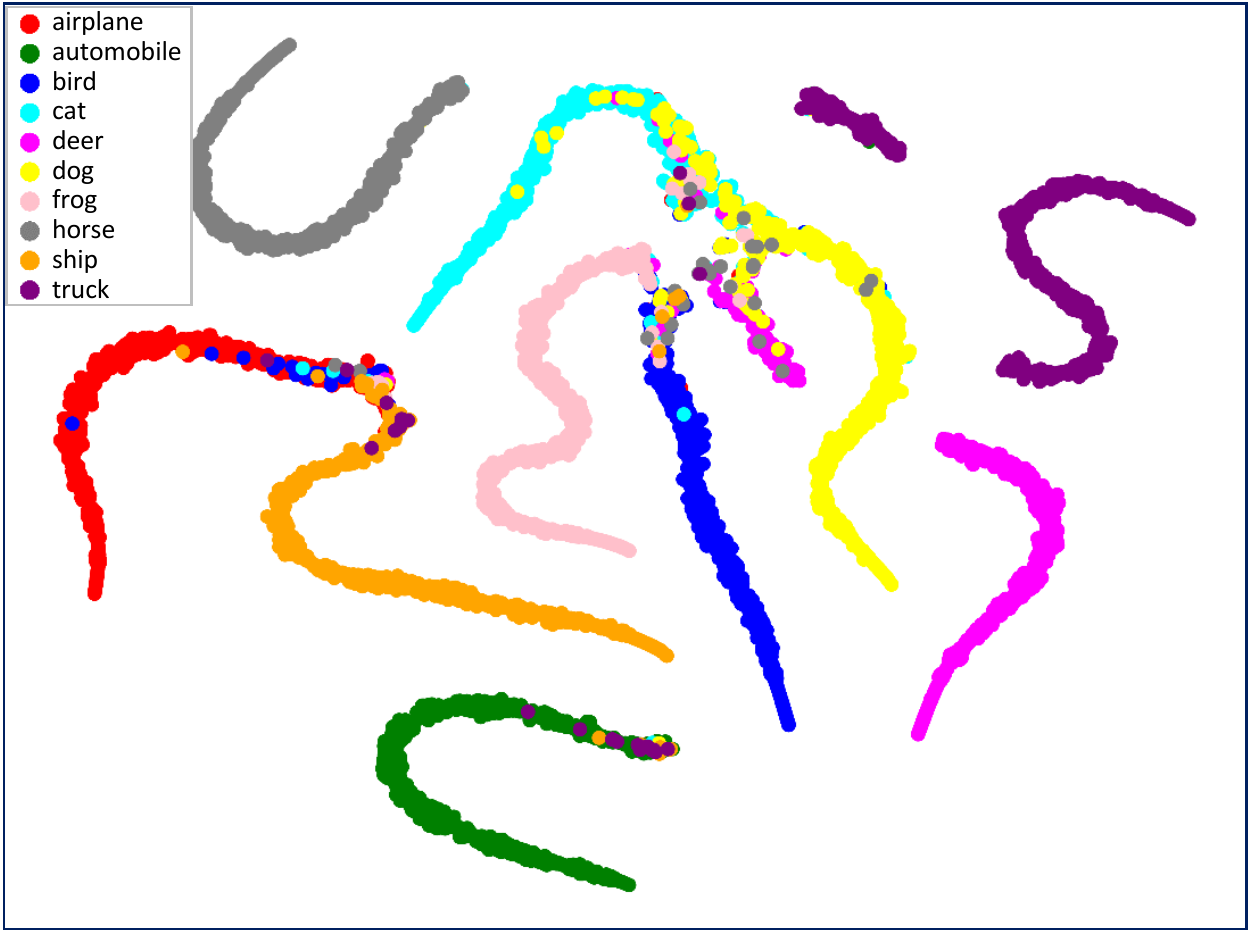}
\label{fig_main_methods_fixmatch}}
\\
\vskip 0.1in
\subfloat[ReRankMatch$^{\text{BM}}$ on SVHN]{\includegraphics[width=2.95in]{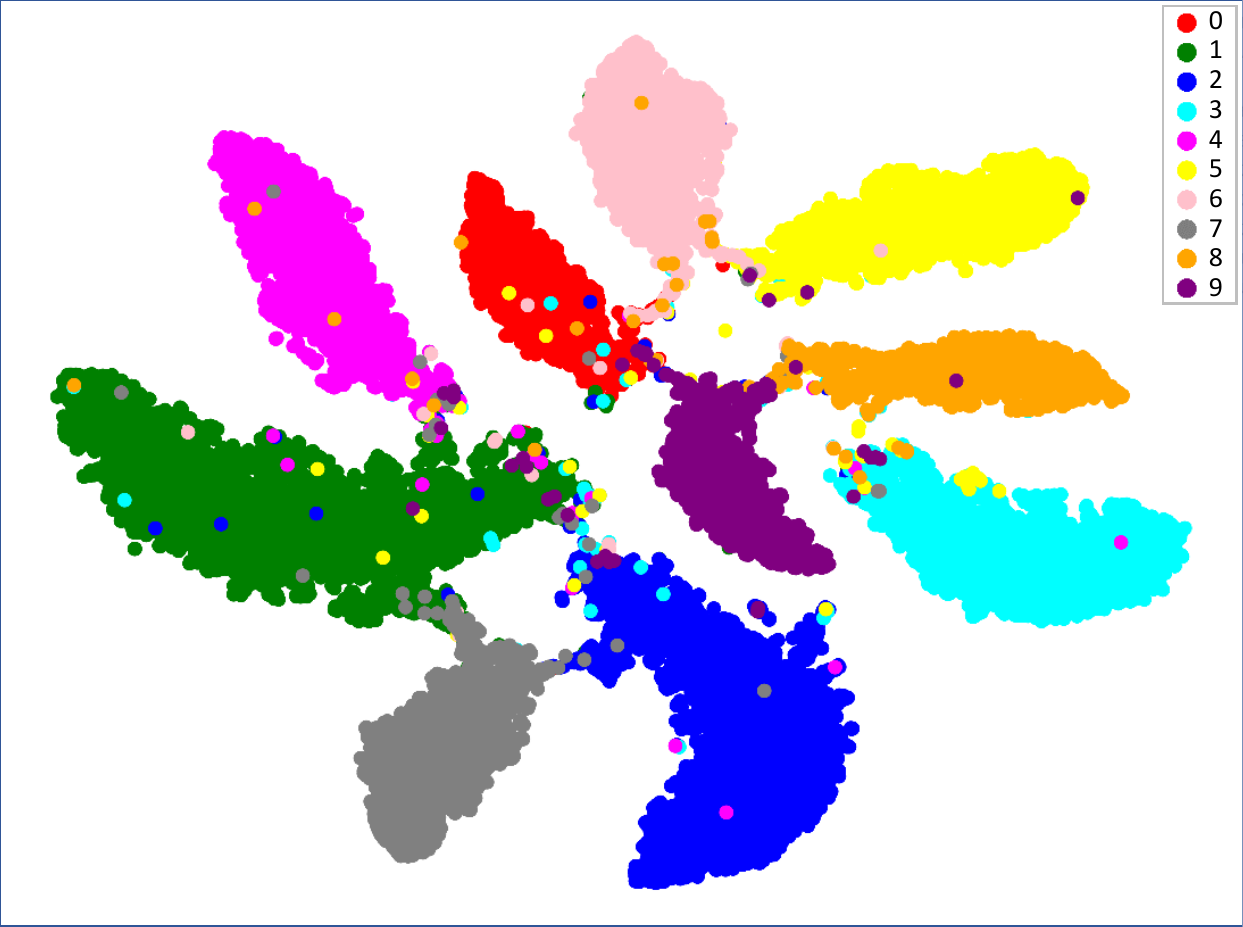}
\label{fig_main_methods_rankingbm}}
\hfil
\subfloat[ReRankMatch$^{\text{CT}}$ on SVHN]{\includegraphics[width=2.95in]{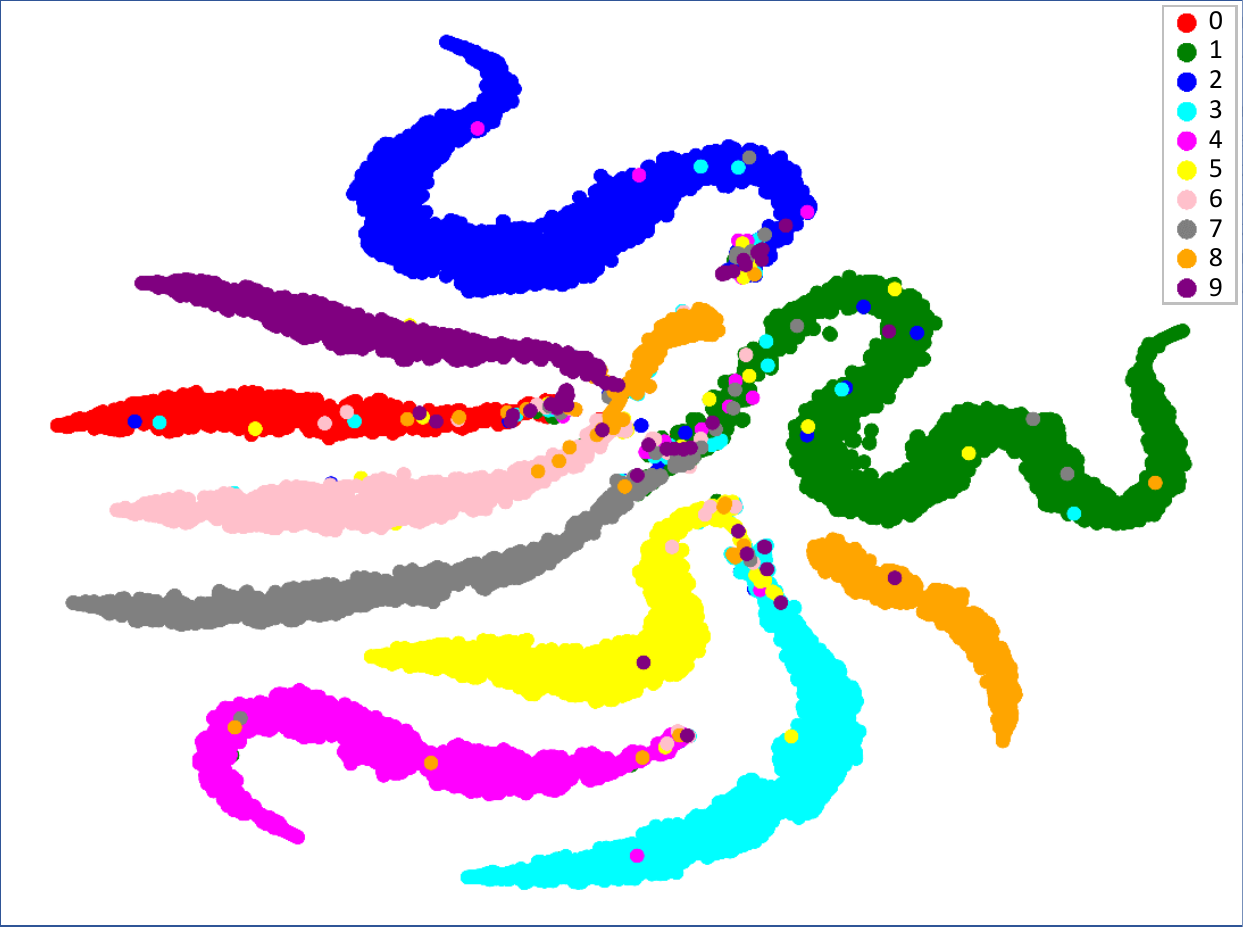}
\label{fig_main_methods_rankingct}}
\caption{t-SNE visualization of our method on CIFAR-10 and SVHN test set. (a)(b) The models were trained for 256 epochs with 4000 labels. (c)(d) The models were trained for 128 epochs with 1000 labels. The same color depicts the same class.}
\label{fig_vis}
\end{figure*}

\subsubsection{Threshold and Margin in Semantics-Oriented Part}

\begin{table}[h!]
\centering
\scalebox{1.1}{
\begin{tabular}{@{\hskip 0.01in}l@{\hskip 0.11in}rrr}
\toprule
\shortstack[l]{Ablation\\ReRankMatch$^{\text{CT}}$}            & \shortstack[r]{CIFAR-10\\4000 labels} & \shortstack[r]{SVHN\\1000 labels} & \shortstack[r]{Tiny ImageNet\\9000 labels} \\
\midrule
$\psi = 0.5$, $\phi = 0.3$ & 95.68 & 97.90 & 45.17 \\
$\psi = 0.5$, $\phi = 0.5$ & 95.79 & 97.81 & 45.06 \\
$\psi = 0.5$, $\phi = 1.0$ & 95.70 & 97.84 & 45.51 \\
$\psi = 0.3$, $\phi = 0.3$ & 95.68 & 97.77 & 45.31 \\
$\psi = 0.1$, $\phi = 0.3$ & 95.69 & 97.85 & 45.96 \\
\bottomrule
\end{tabular}
}
\caption{Ablation study with threshold $\psi$ and margin $\phi$. All results are top-1 accuracy (\%).}
\label{table_ablation}
\end{table}

We conduct an ablation study with various values of the threshold $\psi$ and margin $\phi$, as shown in Table \ref{table_ablation}. The meaning of $\psi$ and $\phi$ was presented in Section \ref{sec_semantics_oriented}. Our main results were reported with a threshold $\psi$ of 0.5 and a margin $\phi$ of 0.3. We first keep the threshold $\psi$ unchanged and increase the margin $\phi$, as reported in the $2^{nd}$ and $3^{rd}$ row of Table \ref{table_ablation}. We then decrease the threshold $\psi$ while keeping the margin $\phi$ as a constant, as shown in the $4^{th}$ and $5^{th}$ row of Table \ref{table_ablation}. The results tend to be persistent across the cases and datasets, implying that the performance does not strictly depend on threshold and margin values.

\subsection{Qualitative Results}

To cast the light for how our models have learned to classify the images, we visualize the ``logits" scores using t-SNE introduced by \cite{van2008visualizing}. t-SNE visualization reduces the high-dimensional features to a reasonable dimension to help grasp the tendency of the learned models. Figure \ref{fig_vis} shows t-SNE visualization of the ``logits" scores of ReRankMatch on CIFAR-10 and SVHN. ReRankMatch obtains a good classification when the same-class samples are grouped into the same cluster. The visualization also shows that the boundaries between clusters are clearly distinguishable, though there are still overlapping points among classes. Interestingly, there is a significant difference in the shape of the clusters between ReRankMatch$^{\text{BM}}$ and ReRankMatch$^{\text{CT}}$. This difference might be due to the function used to compute the error between $L_2$-normalized ``logits" scores: ReRankMatch$^{\text{BM}}$ uses Euclidean distance, while ReRankMatch$^{\text{CT}}$ utilizes cosine similarity.

\subsection{Analysis of Training Loss Curve}

\begin{figure}[h!]
\centering
\includegraphics[width=3.4in]{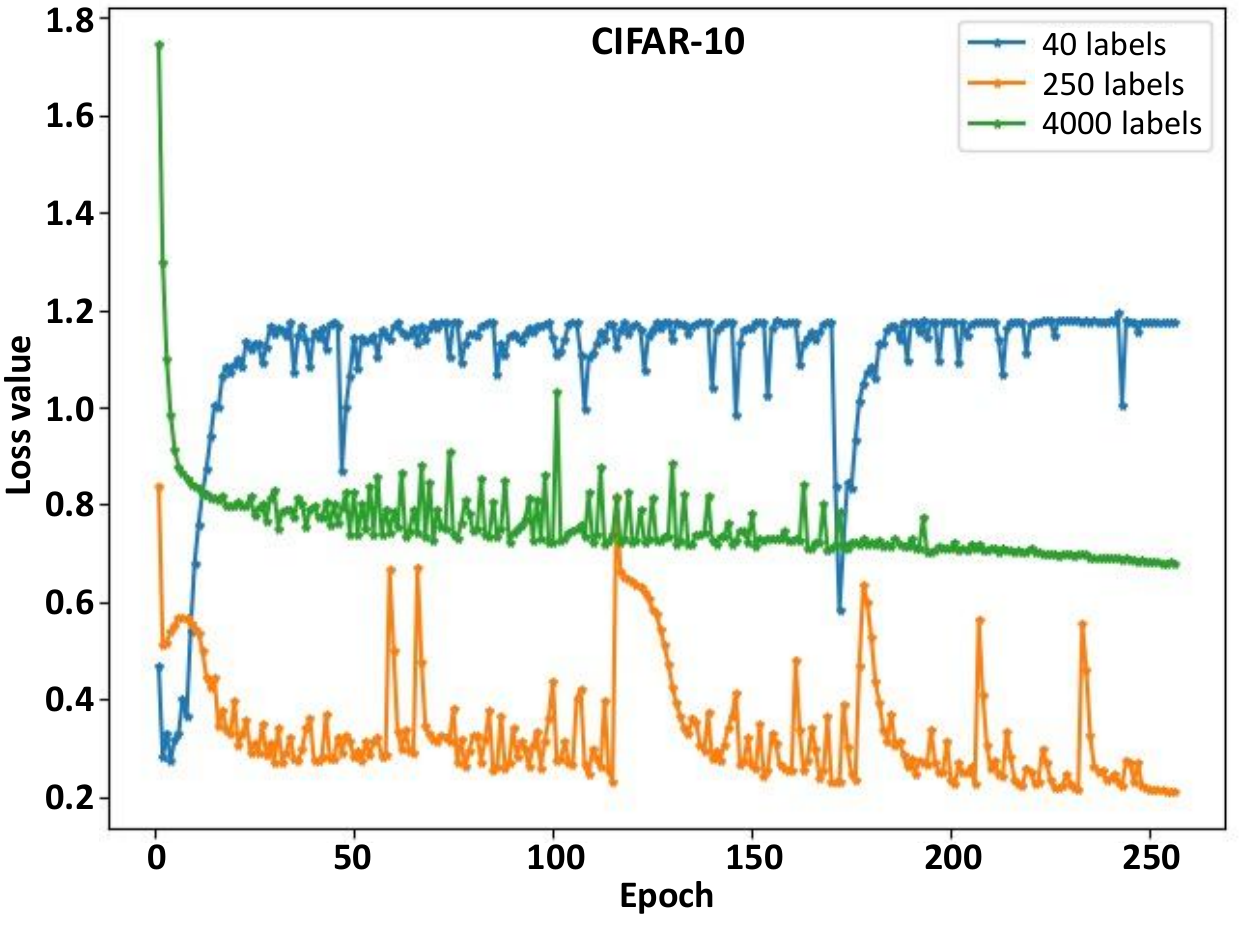}
\caption{Training loss curve of ReRankMatch$^{\text{BM}}$.}
\label{fig_loss_curve}
\end{figure}

Training loss curves of ReRankMatch$^{\text{BM}}$ trained on CIFAR-10 with 4000, 250, and 40 labels are illustrated in Figure \ref{fig_loss_curve}.

As shown in Figure \ref{fig_loss_curve}, the model converges in the case of 4000 and 250 labels, but the model does not converge with a very small amount of labeled data. Therefore, ReRankMatch needs a sufficient amount of labeled data to learn semantics-oriented similarity representation.

\section{Conclusion}

In this paper, we introduce a novel method, named ReRankMatch, integrating semantics-oriented similarity representation into the recently proposed RankingMatch. Our method aims to deal with the case in which labeled and unlabeled data share non-overlapping categories. ReRankMatch consists of two main parts: class-specific and semantics-oriented. While the class-specific part encourages the model to produce the similar ``logits" scores for not only the different perturbations of the same input but also the same-class inputs, the semantics-oriented part leads the model to produce the similar image representations for the inputs likely belonging to the same class. We conduct extensive experiments on various datasets to verify the usefulness of semantics-oriented similarity representation. Our results show that semantics-oriented similarity representation is meaningful and helpful if there is sufficient amount of labeled data. Especially, ReRankMatch achieves state-of-the-art results on several datasets such as CIFAR-10, CIFAR-100, and SVHN. For future work, we are going to do more studies to improve the performance of semantics-oriented similarity representation when the labeled data is scanty.

\section*{Acknowledgment}
This research was supported by the MSIT(Ministry of Science and ICT), Korea, under the Grand Information Technology Research Center support program(IITP-2021-2020-0-01489) supervised by the IITP(Institute for Information \& communications Technology Planning \& Evaluation), and the Technology Innovation Program(or Industrial Strategic Technology development Program, 2000682, Development of Automated Driving Systems and Evaluation) funded by the Ministry of Trade, Industry \& Energy(MOTIE, Korea).

\bibliography{egbib}

\begin{thebibliography}{10}
\providecommand{\url}[1]{#1}
\csname url@samestyle\endcsname
\providecommand{\newblock}{\relax}
\providecommand{\bibinfo}[2]{#2}
\providecommand{\BIBentrySTDinterwordspacing}{\spaceskip=0pt\relax}
\providecommand{\BIBentryALTinterwordstretchfactor}{4}
\providecommand{\BIBentryALTinterwordspacing}{\spaceskip=\fontdimen2\font plus
\BIBentryALTinterwordstretchfactor\fontdimen3\font minus
  \fontdimen4\font\relax}
\providecommand{\BIBforeignlanguage}[2]{{%
\expandafter\ifx\csname l@#1\endcsname\relax
\typeout{** WARNING: IEEEtran.bst: No hyphenation pattern has been}%
\typeout{** loaded for the language `#1'. Using the pattern for}%
\typeout{** the default language instead.}%
\else
\language=\csname l@#1\endcsname
\fi
#2}}
\providecommand{\BIBdecl}{\relax}
\BIBdecl

\bibitem{kolesnikov2019big}
A.~Kolesnikov, L.~Beyer, X.~Zhai, J.~Puigcerver, J.~Yung, S.~Gelly, and
  N.~Houlsby, ``Big transfer (bit): General visual representation learning,''
  \emph{arXiv preprint arXiv:1912.11370}, vol.~6, p.~3, 2019.

\bibitem{xie2020self}
Q.~Xie, M.-T. Luong, E.~Hovy, and Q.~V. Le, ``Self-training with noisy student
  improves imagenet classification,'' in \emph{Proceedings of the IEEE/CVF
  Conference on Computer Vision and Pattern Recognition}, 2020, pp.
  10\,687--10\,698.

\bibitem{clark2017simple}
C.~Clark and M.~Gardner, ``Simple and effective multi-paragraph reading
  comprehension,'' \emph{arXiv preprint arXiv:1710.10723}, 2017.

\bibitem{noroozi2016unsupervised}
M.~Noroozi and P.~Favaro, ``Unsupervised learning of visual representations by
  solving jigsaw puzzles,'' in \emph{European Conference on Computer
  Vision}.\hskip 1em plus 0.5em minus 0.4em\relax Springer, 2016, pp. 69--84.

\bibitem{gidaris2018unsupervised}
S.~Gidaris, P.~Singh, and N.~Komodakis, ``Unsupervised representation learning
  by predicting image rotations,'' \emph{arXiv preprint arXiv:1803.07728},
  2018.

\bibitem{berthelot2019mixmatch}
D.~Berthelot, N.~Carlini, I.~Goodfellow, N.~Papernot, A.~Oliver, and C.~A.
  Raffel, ``Mixmatch: A holistic approach to semi-supervised learning,'' in
  \emph{Advances in Neural Information Processing Systems}, 2019, pp.
  5049--5059.

\bibitem{sohn2020fixmatch}
K.~Sohn, D.~Berthelot, C.-L. Li, Z.~Zhang, N.~Carlini, E.~D. Cubuk, A.~Kurakin,
  H.~Zhang, and C.~Raffel, ``Fixmatch: Simplifying semi-supervised learning
  with consistency and confidence,'' \emph{arXiv preprint arXiv:2001.07685},
  2020.

\bibitem{schroff2015facenet}
F.~Schroff, D.~Kalenichenko, and J.~Philbin, ``Facenet: A unified embedding for
  face recognition and clustering,'' in \emph{Proceedings of the IEEE
  conference on computer vision and pattern recognition}, 2015, pp. 815--823.

\bibitem{hermans2017defense}
A.~Hermans, L.~Beyer, and B.~Leibe, ``In defense of the triplet loss for person
  re-identification,'' \emph{arXiv preprint arXiv:1703.07737}, 2017.

\bibitem{anonymous2021rankingmatch}
\BIBentryALTinterwordspacing
T.~Q. Tran, M.~Kang, and D.~Kim, ``Rankingmatch: Delving into semi-supervised
  learning with consistency regularization and ranking loss,'' 2021. [Online].
  Available: \url{https://openreview.net/forum?id=T1EMbxGNEJC}
\BIBentrySTDinterwordspacing

\bibitem{chen2020learning}
Y.-C. Chen, C.-T. Chou, and Y.-C.~F. Wang, ``Learning to learn in a
  semi-supervised fashion,'' \emph{arXiv preprint arXiv:2008.11203}, 2020.

\bibitem{cubuk2020randaugment}
E.~D. Cubuk, B.~Zoph, J.~Shlens, and Q.~V. Le, ``Randaugment: Practical
  automated data augmentation with a reduced search space,'' in
  \emph{Proceedings of the IEEE/CVF Conference on Computer Vision and Pattern
  Recognition Workshops}, 2020, pp. 702--703.

\bibitem{devries2017improved}
T.~DeVries and G.~W. Taylor, ``Improved regularization of convolutional neural
  networks with cutout,'' \emph{arXiv preprint arXiv:1708.04552}, 2017.

\bibitem{nair2019realmix}
V.~Nair, J.~F. Alonso, and T.~Beltramelli, ``Realmix: Towards realistic
  semi-supervised deep learning algorithms,'' \emph{arXiv preprint
  arXiv:1912.08766}, 2019.

\bibitem{berthelot2019remixmatch}
D.~Berthelot, N.~Carlini, E.~D. Cubuk, A.~Kurakin, K.~Sohn, H.~Zhang, and
  C.~Raffel, ``Remixmatch: Semi-supervised learning with distribution matching
  and augmentation anchoring,'' in \emph{International Conference on Learning
  Representations}, 2019.

\bibitem{krizhevsky2009learning}
A.~Krizhevsky, G.~Hinton \emph{et~al.}, ``Learning multiple layers of features
  from tiny images,'' 2009.

\bibitem{netzer2011reading}
Y.~Netzer, T.~Wang, A.~Coates, A.~Bissacco, B.~Wu, and A.~Y. Ng, ``Reading
  digits in natural images with unsupervised feature learning,'' 2011.

\bibitem{coates2011analysis}
A.~Coates, A.~Ng, and H.~Lee, ``An analysis of single-layer networks in
  unsupervised feature learning,'' in \emph{Proceedings of the fourteenth
  international conference on artificial intelligence and statistics}, 2011,
  pp. 215--223.

\bibitem{zagoruyko2016wide}
S.~Zagoruyko and N.~Komodakis, ``Wide residual networks,'' \emph{arXiv preprint
  arXiv:1605.07146}, 2016.

\bibitem{loshchilov2016sgdr}
I.~Loshchilov and F.~Hutter, ``Sgdr: Stochastic gradient descent with warm
  restarts,'' \emph{arXiv preprint arXiv:1608.03983}, 2016.

\bibitem{oliver2018realistic}
A.~Oliver, A.~Odena, C.~A. Raffel, E.~D. Cubuk, and I.~Goodfellow, ``Realistic
  evaluation of deep semi-supervised learning algorithms,'' in \emph{Advances
  in neural information processing systems}, 2018, pp. 3235--3246.

\bibitem{zhao2015stacked}
J.~Zhao, M.~Mathieu, R.~Goroshin, and Y.~Lecun, ``Stacked what-where
  auto-encoders,'' \emph{arXiv preprint arXiv:1506.02351}, 2015.

\bibitem{denton2016semi}
E.~Denton, S.~Gross, and R.~Fergus, ``Semi-supervised learning with
  context-conditional generative adversarial networks,'' \emph{arXiv preprint
  arXiv:1611.06430}, 2016.

\bibitem{van2008visualizing}
L.~Van~der Maaten and G.~Hinton, ``Visualizing data using t-sne.''
  \emph{Journal of machine learning research}, vol.~9, no.~11, 2008.

\end{thebibliography}
\end{document}